\documentclass[10pt]{article}
\usepackage[letterpaper]{geometry}
\usepackage{hicss}
\usepackage{times}
\usepackage[none]{hyphenat}
\usepackage{url}
\usepackage{latexsym}
\usepackage{indentfirst}
\usepackage{graphicx}
\graphicspath{{images/}}

\usepackage{graphicx}
\usepackage{color}
\usepackage{booktabs}
\usepackage{caption} 
\usepackage[subrefformat=parens]{subcaption}
\usepackage{amsmath}
\usepackage{url}
\usepackage{algorithm,algorithmic}
\usepackage{bm}
\usepackage{colortbl}

\usepackage {framed,color}

\newcommand{\RED}[1]{#1}
\newcommand{\RRED}[1]{#1}

\usepackage{ascmac}
\def\Underline{\setbox0\hbox\bgroup\let\\\endUnderline}
\def\endUnderline{\vphantom{y}\egroup\smash{\underline{\box0}}\\}
\def\|{\verb|}

\allowdisplaybreaks[4]

\title{Evacuation Shelter Scheduling Problem}

\author{Hitoshi Shimizu \\
 NAIST, NTT, Japan \\
 {\underline{hitoshi.shimizu.kg@hco.ntt.co.jp}} \\
\\
Akinori Fujino \\
 NTT, Japan\\
 {\underline{akinori.fujino.yh@hco.ntt.co.jp}} \\
 \And 
 Hirohiko Suwa \\
 NAIST, Japan \\
 {\underline{h-suwa@is.naist.jp}} \\
\\
 Hiroshi Sawada \\
 NTT, Japan \\
 {\underline{hiroshi.sawada.wn@hco.ntt.co.jp}}\\
 \And 
 Tomoharu Iwata \\
 NTT, Japan \\
 {\underline{tomoharu.iwata.gy@hco.ntt.co.jp}}\\
\\
 Keiichi Yasumoto \\
 NAIST, Japan \\
 {\underline{yasumoto@is.naist.jp}}
}

\date{}

\begin{document}
\maketitle
\begin{abstract}
Evacuation shelters, which are urgently required during natural disasters,
are designed to minimize the burden of evacuation on human survivors.
However, the larger the scale of the disaster, the more costly it becomes to operate shelters.
When the number of evacuees decreases, 
the operation costs can be reduced by moving the remaining evacuees to other shelters and closing shelters as quickly as possible.
On the other hand, relocation between shelters imposes a huge emotional burden on evacuees.
In this study, we formulate the ``Evacuation Shelter Scheduling Problem,'' which
allocates evacuees to shelters in such a way
to minimize the movement costs of the evacuees and the operation costs of the shelters.
Since it is difficult to solve this quadratic programming problem directly,
we show its transformation into a 0-1 integer programming problem.
In addition, such a formulation struggles to calculate the burden of relocating them from historical data  because no payments are actually made.
To solve this issue,
we propose a method that estimates movement costs based on the numbers of evacuees and shelters
during an actual disaster.
Simulation experiments with records from the Kobe earthquake
(Great Hanshin-Awaji Earthquake) showed
that our proposed method reduced operation costs by 33.7 million dollars: 32\%.

\end{abstract}

\section{Introduction}


\urldef\easturl\url{https://en.wikipedia.org/wiki/2011_T%C5%8Dhoku_earthquake_and_tsunami}

During natural disasters, evacuation shelters provide temporary lodging and safety to survivors.
Since the Great East Japan Earthquake (March 2011)\footnote{\easturl (accessed June 14, 2021)}
forced approximately 470,000 people to evacuate their homes,
more evacuation shelters have been set up in various cities.
For example, the Tokyo Metropolitan Government established guidelines to develop evacuation shelters within a 3-km distance of evacuated areas.%
\footnote{
\url{https://www.bousai.metro.tokyo.lg.jp/bousai/1000026/1000316.html}
(accessed June 14, 2021)
}
Such government guidelines will 
increase the number of evacuation shelters
and reduce the evacuation burden on evacuees.

However, operating shelters also brings a burden.
For example, when a school is used as an shelter, 
educational activities will be hindered if evacuees are staying there
when classes are resumed after the disaster.
As evacuees remain longer in a shelter,
the slower they will recover from the disaster.
The burden of shelter operation can be translated into monetary terms.
Otsuka et al.~\cite{otsuka2016} estimated that,
based on the rental expenses of the facilities occupied during the Kobe earthquake
(Great Hanshin-Awaji Earthquake)%
\footnote{
On January 17, 1995, a major earthquake of magnitude 6.9 occurred in southern Hyogo Prefecture, Japan.
Buildings were severely damaged, mainly in Kobe City,
and about 300 fires broke out.
A total of 6,434 people were killed.
\url{https://en.wikipedia.org/wiki/Great_Hanshin_earthquake}
(accessed June 14, 2021)
}
the operation cost of its shelters was about 106.6 million dollars (10.66 billion yen).%
\footnote{
We used the exchange rate in January 1995: 100 yen to one dollar.
}
In contrast, cases can also be found where the number of evacuees is small compared to the number of shelters.%
\footnote{
For example, in the earthquake off the coast of Fukushima Prefecture on February 13, 2021,
no fatalities were confirmed,
70 evacuation centers were opened,
and 120 people were evacuated.
\url{https://www.asahi.com/articles/ASP2F7XHXP2FDIFI00D.html}
(accessed June 14, 2021)
}
In such cases, it might have been possible to gather the evacuees into fewer shelters to save on operational costs.

In this paper, we discuss possible conflicts between the demands to reduce the evacuation cost of evacuees and the operation cost of shelters.
The facility location problem (FLP)~\cite{cornuejols1983uncapicitated, daskin2008you}
has been studied to minimize total movement and operation costs.
\RED{
FLP is one of the fundamental combinatorial optimization problems
 in the field of location intelligence,}
which determines the location of facilities from candidate sites and
the allocation of users and facilities to satisfy users' demand.
However, since the FLP does not take time into account, it cannot represent a situation in which the number of evacuees changes.
For example, when the number of evacuees decreases, operation costs can be lowered by relocating the remaining evacuees to another shelter and closing shelters as quickly as possible.
\RED{
On the other hand,
Nakahira~\cite{Nakahira2018} reported that
the relocation of evacuees was difficult because of
not only the burden of leaving the accustomed shelters close to their homes,
but the fact that some of them were not cooperative
as they were treated as obstacles to the resumption of school classes.
}

To find an acceptable solution to both evacuees and other residents under these circumstances,
we must consider not only the initial cost of opening shelters during a disaster but also the cumulative cost until all of them are closed.
In this study, we extend the FLP in time and formulate an allocation problem that minimizes
the sum of the operation cost of shelters and the movement cost of evacuees until all return home.
By solving this problem, we can determine which shelters should be closed 
and to which shelters should the remaining evacuees be relocated to minimize the costs
when the evacuees gradually return home as their number decreases.
\RED{
In addition, the analysis of the solutions can provide useful insights for determining the location and scale of new shelters.}

However, determining the burden of relocating evacuees from historical data is complicated
because it is not actually paid.
We solve this problem by
proposing a method to estimate the movement cost of evacuees
based on the number of evacuees and shelters during an actual disaster.
At the beginning of the proposed method, 
the movement cost parameters are estimated by assuming the sequential FLP’s allocation ({\sc SeqFlp}),
which is a procedure for sequentially finding the solution of FLP.
Then using the obtained parameters, the OPTimal allocation ({\sc Opt})  is executed,
which assigns the evacuees to shelters to minimize the accumulative costs of 
both the movement of the evacuees and the operation of the shelters.
To evaluate how effectively the proposed method reduced operation costs,
we conducted a simulation experiment in which the conditions were set based on the records of the Kobe earthquake.

\RRED{
Our formulation will not only improve the efficiency in the recovery phase of disaster management,
but expand the scope of the scheduling problem using combinatorial optimization.
These are contributions to the future development of location intelligence.
}


\section{Related Work}
\RED{
\subsection{Evacuation Shelter Allocation}

The problem of shelter allocation has been studied in various aspects of disaster operations management (DOM),
including the types of disaster, the means of evacuation,
the types of objective functions and constraints,
and the types of contributions
(theoretical or applied)~\cite{amideo2019optimising}.
For example, the suitability of evacuation centers~\cite{kar2008gis}, 
the fire risk of evacuation shelters and evacuation routes~\cite{alccada2009multiobjective,coutinho2012solving},
and the convenience for pets and the elderly~\cite{kocatepe2018pet} have been evaluated as objects of disaster countermeasures.
However, \cite{galindo2013review,altay2006or} argue that
research on the recovery phases in disasters,
such as our study, is lacking and desired compared to
phases of mitigation, preparedness, and response.


Similar to our study, various formulations of integer programming problems have been proposed.
Sherali et al.~\cite{sherali1991location} formulated the evacuation planning model (EPM) as a nonlinear mixed integer programming problem and proposed both exact and heuristic methods.
The objective is to minimize the evacuation time so as to satisfy the capacity constraint of shelters.
However, the operation cost of shelters was not incorporated into the objective function.
Swamy et al.~\cite{swamy2017hurricane} redefined the Bus Evacuation Problem (BEP) in a planning framework,
and simulated escaping from a hurricane in New York City.
Chen et al.~\cite{chen2013temporal} proposes a three-stage hierarchical location model that takes into account the progress of time, and balances the trade-off between evacuation efficiency and planning budget.
Our approach is most similar to \cite{chen2013temporal},
but it is extended to consider the decrease of the number of evacuees
and to automatically resolve the trade-off between evacuation efficiency and operation cost by estimating the movement cost. 

\subsection{Facility Location Problem}
}
The facility location problem (FLP),
which takes into account the distance from users to determine appropriate locations, 
has been studied for decades and can be traced back to the work of Weber~\cite{weber1929theory} and Hotelling~\cite{hotelling1929stability}.
The range of applications of FLP is wide.
A typical example is the layout of factories and warehouses.
However, it can also be applied to hospitals,
fire stations, schools, waste disposal plants, etc.~\cite{daskin2008you}.
The research in this paper is based on a kind of FLP where the location of users and candidate locations are discrete, and the sum of the cost of the facility and the cost of the combination of facilities and users is minimized.
There are two types of variations among such problems:
the simple type, the uncapacitated facility location problem (UFLP) \cite{robers1976study,daskin2008you,kochetov2011facility},
and the constrained type, the capacitated facility location problem, (CFLP) \cite{wu2006capacitated}.
In the formulation of this study,
the case where $T=1$ coincides with CFLP.

Some temporally extended models of the FLP have also been developed:
The dynamic facility location problem (DFLP) deals with fluctuating demands~\cite{van1982dual,nickel2019multi},
and some studies have examined FLPs in which
factors other than demand behave in a stochastic manner~\cite{owen1998strategic,correia2019facility}.
The multi-stage uncapacitated facility location problem (MSUFLP) was formulated as a model that studies routes that pass through multiple facilities~\cite{kochetov2011facility}.
Some mathematical methods~\cite{takizawa2012emergency,takizawa2013enumeration}
for evacuation shelter location planning have been proposed, as in this study.
However, to the best of our knowledge, no model has been formulated that can investigated the cost of relocating evacuees to nearby shelters.

\section{Problem Formulation}
\label{sec:setting}

In section \ref{form:flp},
we present a formulation of the shelter allocation problem without time evolution.
In section \ref{form:setting},
 we extend it to a problem that addresses time.

\subsection{Facility Location Problem}
\label{form:flp}

Let $N$ be the set of evacuees, and let $M$ be the set of shelters.
Consider a situation where all the evacuees in $N$ are evacuated to
one of the shelters in $M$.
Shelter $m$ can accommodate $C_m$ evacuees at most.
To ensure that all the evacuees are accommodated,
we assume $\sum_{m \in M} C_m \geq |N|$.
Let $d_{mn}$ be the evacuation cost of moving evacuee $n$ to shelter $m$.
Let $f_{m}$ denote the operation cost of running shelter $m$.%
\footnote{
Operation costs is assumed to be independent of the number of evacuees to be accommodated 
since they correspond to the opportunity cost of not being able to use the facility for its original purpose.
}
Here we introduce variable $x_{mn}$, which indicates whether evacuee $n$ will be accommodated in shelter $m$ or not,
and variable $y_{m}$, which indicates whether shelter $m$ will be operated or not.
When we minimize the sum of movement cost $\sum_{m \in M}\sum_{n \in N} d_{mn} x_{mn}$ and operation cost $\sum_{m \in M} f_m y_{m}$, i.e., the following equation (\ref{eq:obj}'),
we find the shelter allocation: variables $x_{mn}$ and $y_{m}$.
A problem of this form is called a facility location problem (FLP)~\cite{cornuejols1983uncapicitated}.
Note that the equation numbers correspond to those of the proposed method:

\begin{align}
& \text{Minimize} & &  \sum_{m \in M} \sum_{n \in N} d_{mn} x_{mn} + \sum_{m \in M} f_m y_{m}  \tag{\ref{eq:obj}'}\\
& \text{Subject to} & & \sum_{n \in N} x_{mn} \leq C_m y_{m} ,  \forall m \tag{\ref{eq:capa}'} \\
&  & & \sum_{m \in M} x_{mn} = 1, \forall n \tag{\ref{eq:alive}'}\\
& & & x_{mn}, y_{m} \in \{ 0,1 \} \tag{\ref{eq:xrange}'}
\end{align}

\subsection{Evacuation Shelter Scheduling Problem}
\label{form:setting}

By solving the problem setting in section \ref{form:flp},
we can find the optimal allocation considering both the costs of evacuation to shelters and operating them.
However, this FLP does not take time into account.
Therefore, 
when the evacuees start to eventually return home and their numbers decrease,
the solution of FLP cannot determine which shelters should be closed to minimize costs.

Therefore, we extend the FLP in time
and assume that evacuees in a shelter can be relocated to another shelter
 to reduce the total number of shelters.
This extended formulation creates a shelter management plan that minimizes the total movement and operation costs.
The symbols in this section are defined in Table~\ref{table:parameters}.
\begin{figure}[tb]
\centering
\includegraphics[width=7cm]{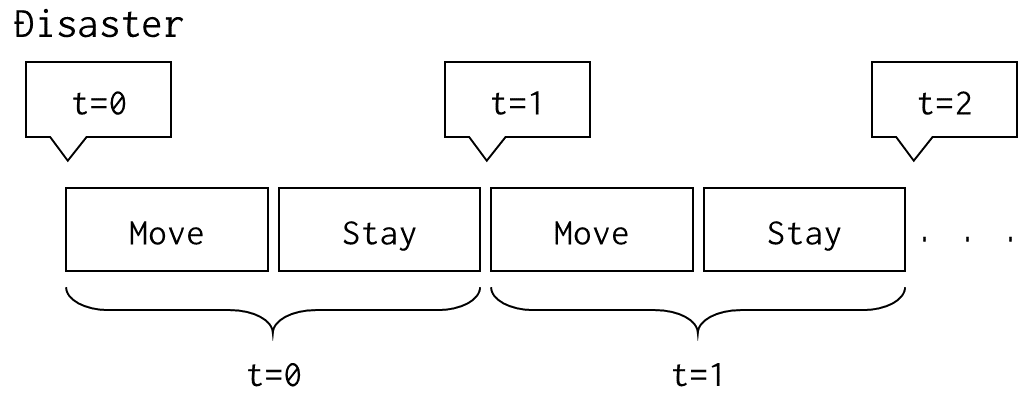}
\caption{Time step in Evacuation Shelter Scheduling Problem}
\label{fig:time}
\end{figure}
\begin{table}[tb]
\small
\centering
\caption{Notation}
\begin{tabular}{ll}
\toprule
Symbols & 
Definitions\\
\midrule
$N$ & 
Set of indexes of evacuees: $n \in \{1, \cdots , |N| \}$\\
$N_t$ & 
Set of indexes of evacuees in a shelter at time $t$\\
$M$ & 
Set of indexes of locations: $m \in \{1, \cdots , |M| \}$ \\
$M_t$ & 
Set of indexes of shelters that can be operated \\
& at time $t$. 
\\
$\tilde{m}_t(n)$ & 
Location of evacuee $n$ at time $t$
\\
$L$ & 
Set of indexes of types of facilities:\\
& $\ell \in \{1, \cdots , |L| \}$ 
\\
$C_m$ & 
Capacity of shelter $m$
\\
$T$ & 
Maximum value of time to be considered:\\
& $t \in \{0, \cdots , T \}$
\\
$\tau_{n}$ & 
Time when evacuee $n$ returns home
\\
$f_{m}$ & 
Operation cost for one time step of shelter $m$.
\\
$d_{tmm'}$ & 
Movement cost of evacuee from location $m$ \\
& at time $t$ to $m'$ at time $t+1$.
\\
$x_{tmn}$ & 
Indicator that the location of evacuee $n$ \\
& at time $t$ is location $m$. \\
$y_{tm}$ & Indicator for operating shelter $m$ at time $t$.\\
$z_{tmm'n}$ &  Indicator for the movement of evacuee $n$ from \\
& location $m$ at time $t$ to $m'$ at time $t+1$. \\
$U_{\ell}$ & 
Number of days that facility type $\ell$ is occupied.\\
$s_{m\ell}$ &  Indicator that shelter $m$ is type $\ell$\\
$r_{m}$ & Ratio of real number of shelters to $|M|$\\
\RED{$r_{n}$} & \RED{Ratio of real number of evacuees to $|N|$}\\
$\alpha$ & Ratio of evacuation cost to relocation cost\\
$\bm{p}_{m}$ & 2D coordinate of location $m$\\
 \bottomrule
\end{tabular}
\label{table:parameters}
\end{table}

As in section \ref{form:flp},
let $N$ be the set of evacuees.
Because the source of movement is either the location of an evacuee at the time of the disaster or the location of a shelter,
let $M$ be the set of both locations to treat them in a unified manner,
and an evacuee is assumed to be at one of the locations $M$ during the disaster.
Evacuee $n$ is assumed to 
return home after staying for $\tau_n$ steps.
In this paper, $\tau_n$ is referred to as the return time.%
\footnote{Although $\tau_n$ is generally difficult to know immediately after a disaster,
we assume that it is a known value here.}
Let $T = \max_n \tau_n$.
The time is expressed as discretized integer $t \in \{0, \dots, T \}$
where $t=0$ is the beginning of the disaster, as in Fig.~\ref{fig:time}.
After a disaster occurs at $t=0$,
evacuees move to a shelter and stay there by $t=1$.
After that, evacuee $n$ with $\tau_n=1$ will go home,
and the others, if necessary, will move to another shelter and stay there by $t=2$.
These procedures are assumed to be repeated.
The location of evacuee $n$ at time $t$ 
is denoted by $\tilde{m}_t(n)$.
If an evacuation shelter can be established at location $m$,
it can accommodate $C_m$ evacuees at most.
Otherwise, $C_m=0$ is set for locations that are not candidates for evacuation shelters.
$\sum_{m \in M} C_m \geq |N|$ is assumed so that all the evacuees can be accommodated in the evacuation shelters.

The movement from location $m$ to $m'$ at time $t$ incurs movement cost $d_{tmm'}$,
and the movement between any two locations is assumed to be completed in one time step.
When shelter $m$ is opened, it costs $f_{m}$ per step to operate it.
Since evacuation cost $d_{t=0,mm'}$ immediately after a disaster is assumed to be different from relocation cost $d_{tmm'}, \forall t > 0$,
 movement cost $d_{tmm'}$ is assumed to depend on time $t$.
In the following, we refer to the movement cost at $t=0$ as the evacuation cost,
and the movement cost at $t>0$ as the relocation cost.
Here, as in the FLP,
we introduce variable $x_{tmn}$, which indicates whether evacuee $n$ will be accommodated in shelter $m$ at time $t$ or not,
 and variable $y_{tm}$, which indicates whether shelter $m$ will be operated at time $t$ or not.

In the above setting,
the problem 
can be expressed 
as a 0-1 integer quadratic programming (Binary Quadratic Programming: BQP) problem~\cite{ito2017optimization}.
We call this the ``Evacuation Shelter Scheduling Problem.''

\begin{oframed}
\begin{align}
& \text{Minimize} & & \sum_{t=0}^{T-1} \Big( \sum_{m \in M} \sum_{m' \in M} d_{tmm'} \sum_{n \in N} x_{tmn} x_{(t+1)m'n} \nonumber \\
& & &  \hspace{2cm} + \sum_{m \in M} f_{m} y_{tm} \Big) \label{eq:obj}\\
& \text{Subject to} & & \sum_{n \in N} x_{tmn} \leq C_m y_{tm} ,  \forall m \label{eq:capa}\\
& & & \sum_{m \in M} x_{tmn} = 1, \forall t \leq \tau_n \label{eq:alive}\\
& & & \sum_{m \in M} x_{tmn} = 0, \forall t > \tau_n \label{eq:dead}\\
& & & x_{t=0, mn} = \begin{cases}
 1, &  m = \tilde{m}_0(n)\\
 0, & m \neq \tilde{m}_0(n)
 \end{cases} \label{eq:init}\\
& & & y_{(t+1)m} \leq y_{tm} \label{eq:noreopen} \\ %
& & & x_{tmn}, y_{tm}  \in \{ 0,1 \} \label{eq:xrange}
\end{align}
\end{oframed}
Equation (\ref{eq:obj}) is an objective function that minimizes the sum of the costs of moving the evacuees and operating the shelters.
Eq. (\ref{eq:capa}) is a condition under which no evacuees can stay in the closed shelters
and the number of evacuees in the open shelters does not exceed the capacity.
However, at the time of the disaster ($t=0$), this constraint is not applied
because the evacuees have not yet been accommodated in the shelters.
Eqs. (\ref{eq:alive}) and (\ref{eq:dead}) are the conditions under which the evacuees live in a shelter until they return home.
Eq. (\ref{eq:init}) is the condition where evacuee $n$ is at a given location $\tilde{m}_0(n)$ when the disaster occurs.
Eq. (\ref{eq:noreopen}) denotes a condition where once a shelter is closed, it will not be reopened.
We consider that it is difficult to requisition facilities again once they have resumed their original use.

\RED{
This BQP can be solved by a commercial solver, 
but as shown in the section~\ref{sec:performance},
the performance is poor while the computation time is long.
Therefore, we transformed it}
 into a 0-1 integer linear programming problem~\cite{ito2017optimization}
 by introducing variable $z_{tmm'n}$, which indicates whether evacuee $n$ moves from location $m$ at $t$ to $m'$ at time $t+1$ or not.
The objective function can be rewritten to Eq.~(\ref{eq:obj}')
by modifying Eq.~(\ref{eq:xrange}) into Eq.~({\ref{eq:xrange}'})
and adding Eqs.~(\ref{eq:z1}), (\ref{eq:z2}), and (\ref{eq:z3}) that make 
$z_{tmm'n} = x_{tmn} \times x_{(t+1)m'n}$ as follows:


\begin{align}
& \text{Minimize} & & \sum_{t=0}^{T-1} \Big( \sum_{m \in M} \sum_{m' \in M} d_{tmm'} \sum_{n \in N} z_{tmm'n}  \nonumber \\ 
& & & \hspace{2cm}  + \sum_{m \in M} f_{m} y_{tm} \Big)
\tag{\ref{eq:obj}'}\\ 
& & & x_{tmn}, y_{tm}, z_{tmm'n} \in \{ 0,1 \}  \tag{\ref{eq:xrange}'}\\
& & & z_{tmm'n} \geq x_{tmn} + x_{(t+1)m'n} - 1  \label{eq:z1} \\ %
& & & z_{tmm'n} \leq x_{tmn}  \label{eq:z2} \\ %
& & & z_{tmm'n} \leq x_{(t+1)m'n}.  \label{eq:z3}%
\end{align}



\subsection{Movement Cost Estimation Problem}

The problem setting in section \ref{form:setting} can be solved optimally using an integer linear programming solver by
 appropriately setting the movement and operation costs.
Operation cost $f_m$ can be estimated from the cost of renting the facilities~\cite{otsuka2016}.
However, movement cost $d_{tmm'}$ for evacuees is difficult to determine from historical data because no monetary payments are made.

On the other hand,
the history of the number of evacuees in shelters $|N_t|$~\cite{KobeMinsei1996} 
and the number of occupied days $U_{\ell}$ by facility type $\ell$ has been reported ~\cite{otsuka2016}.
Facility types $\ell$ are categorized by the original use of the shelters,
such as an elementary school and a park (Table~\ref{tab:shelters}).
We introduce variable $s_{m\ell}$, which indicates whether shelter $m$ is type $\ell$ or not.
Then we can calculate the estimated value of $\hat{U}_{\ell}$ from $y_{tm}$ in the solution
by multiplying the ratio $r_m$ of the actual number of shelters to $|M|$:
\begin{align}
\hat{U}_{\ell} = r_m \times \sum_{t=0}^{T-1} \sum_{m \in M} y_{tm} s_{m\ell}.
\label{eq:U}
\end{align}
This $\hat{U}_{\ell}$ can be used to estimate the movement cost.

Since it is still difficult to estimate $d_{tmm'}$ in $T|M|^2$ dimensions, we make the following assumptions:
\begin{align}
d_{0mm'} &= \alpha \lambda \, || \bm{p}_{m} - \bm{p}_{m'}|| \label{eq:move0}\\
d_{tmm'} &= \phantom{\alpha} \lambda \, || \bm{p}_{m} - \bm{p}_{m'}||, \forall t>0, \label{eq:move1}
\end{align}
where $\bm{p}_m$ is the coordinate of location $m$ and 
$||\bm{p}_{m} - \bm{p}_{m'} ||$ is the Euclidean distance
between location $m$ and $m'$.
Then, we set the problem of estimating $\lambda$
 when $N_t, M, \tilde{m}_0(n), C_m, \tau_n, f_m, \bm{p}_m, \alpha, U_{\ell}, s_{ml}, r_m$
 are known.
We call this the ``movement cost estimation problem.''
\RED{
By estimating the movement cost, it is possible to automatically adjust the trade-off between evacuation efficiency and operation cost.
}
\section{Proposed Method}
\label{sec:method}

For the movement cost estimation problem,
we propose the method to estimate $\lambda$
shown in Algorithm~\ref{alg:proposed}.
 \begin{algorithm}[tb]
 \caption{Proposed Method for Movement Cost Estimation Problem}
 \label{alg:proposed}
 \begin{algorithmic}[1]
 \renewcommand{\algorithmicrequire}{\textbf{Input:}}
 \renewcommand{\algorithmicensure}{\textbf{Output:}}
 \REQUIRE Candidate set of $\lambda$, \\
  $N_t, M, \tilde{m}_0(n), C_m, \tau_n, f_m, \bm{p}_m, \alpha$ for estimation,\\
  $U_{\ell}, s_{ml}, r_m$ for evaluaion,
 \ENSURE $\hat{\lambda}$
  \FOR {$\lambda$ in candidate set}
  \STATE Apply {\sc SeqFLP} to the dataset for estimation to obtain $x_{tmn},y_{tm},z_{tmm'n}$.
  \STATE Calculate MSE from obtained $y_{tm}$
  using Eqs.~(\ref{eq:U}) and~(\ref{eq:MSE}) and the dataset for evaluation.
  \ENDFOR
 \STATE Select $\lambda$ with the smallest MSE as $\hat{\lambda}$
 \RETURN $\hat{\lambda}$ 
 \end{algorithmic} 
 \end{algorithm}
First, we assume that past shelter operations were conducted according 
to the procedure {\sc SeqFlp} shown in section \ref{sec:flp}.
Then we select $\lambda$ that best fits observed $U_{\ell}$ in the training dataset.
The mean square error between $\hat{U}_{\ell} $ and $U_{\ell}$,
\begin{align}
 \mathrm{MSE} = \frac{1}{L} \sum_{\ell \in L} (U_{\ell} - \hat{U}_{\ell})^2,
\label{eq:MSE}
\end{align}
 is used as a loss function to evaluate the goodness of the fit of parameter $\lambda$.

Using parameter $\hat{\lambda}$ (hence, movement cost) estimated from the training dataset,
we can find the optimal solution to the problem formulated in section \ref{form:setting}
by an integer linear programming solver.
This procedure is called {\sc Opt}.

\subsection{Sequential FLP model: {\sc SeqFlp}}
\label{sec:flp}

For each time step $ t \in \{ t \mid 0 \leq t \leq T-1 \} $,
we sequentially solve the following FLP:

\newcommand{\argmax}{\mathop{\rm arg~max}\limits}

\begin{align}
& \text{Minimize} & &  \sum_{m \in M_t} \Big( \sum_{n \in N_t} d_{t\tilde{m}_{t}(n)m} x_{tmn} + f_m y_{tm} \Big) \tag{{\ref{eq:obj}}.F}\\
& \text{Subject to} & & \sum_{n \in N_t} x_{tmn} \leq C_m y_{tm} ,  \forall m \in M_t \tag{\ref{eq:capa}.F} \\
&  & & \sum_{m \in M_t} x_{tmn} = 1, \forall n \in N_t \tag{\ref{eq:alive}.F}\\
& & & x_{tmn}, y_{tm} \in \{ 0,1 \}, \tag{\ref{eq:xrange}.F} \\
&\text{where}&& N_t = \{ n \in N \mid \tau_n > t\} \\
& & & M_0 = M \\  
& & & M_{t+1} = \{ m \in M \mid y_{tm} = 1 \} \\
& & & \tilde{m}_{t+1}(n)= \argmax_m (x_{tmn}).
\end{align}

This {\sc SeqFlp} procedure operates the shelters 
by guiding the evacuees to minimize the sum of the operation 
 and movement costs at each time when the return time of the evacuees is unknown.

\section{Experiments}

\subsection{Summary}

In an evaluation experiment, 
we evaluated the effectiveness of our proposed method in a setting that resembled an actual disaster by creating a dataset 
based on the data of Kobe City during the Kobe earthquake~\cite{baba1996great}.
First, we prepared two datasets: a training dataset for estimating the movement cost of evacuees ({\sc Hanshin Train}) 
and a test dataset for evaluating the effectiveness of the proposed method ({\sc Hanshin Test}).
In an experiment with the former, we estimated $\lambda$ 
with the {\sc SeqFlp} model that was fit to real-world data.
Then by applying the parameter to the latter,
 we show that our proposed method {\sc Opt} has the smallest objective function
 among the other methods.

\begin{figure}
\centering
\includegraphics[width=7cm]{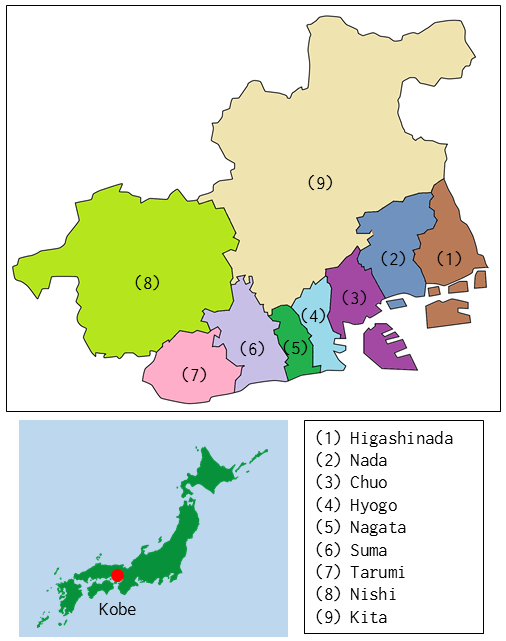}
\caption{Nine wards of Kobe City, all of which were severely damaged by Kobe earthquake.
}
\label{fig:kobe}
\end{figure}

Kobe City’s nine wards are shown in Fig.~\ref{fig:kobe}.
For each one,
we set the number of shelters and evacuees who stayed there in Tables~\ref{tab:shelters} and \ref{tab:evacuee}.
We assume that $\lambda$ is identical for all the wards.

\subsection{Dataset: {\sc Hanshin}}
\label{sec:dataset}
The number of shelters and their operation costs were set (Table~\ref{tab:shelters}), based on two previous works \cite{otsuka2016, kumagai1995}.
Since we couldn’t identify from the previous literature the practical capacity of the shelters despite the legal capacity,
we set a value proportional to the average area~\cite{otsuka2016} of the facilities that were used.
The numbers of the evacuees and their return times were set
to decrease in proportion to the actual number of people staying
in the shelters of the Kobe earthquake~\cite{KobeMinsei1996,kumagai1995} (Table~\ref{tab:evacuee}).
We also set $T=8$ and time step to one month:
covering from January 17, when the earthquake occurred, 
to August 20, when Kobe City closed all the evacuation shelters.
The evacuees were accommodated in a shelter of the ward they were at $t=0$,
and no cross-district movement was allowed.
Since our proposed method cannot be applied with the actual number of shelters and evacuees due to an excessive amount of calculations,
we set the total numbers of evacuees to 1000
and the shelters to 100 for all of Kobe City.
\RED{
However, since both are set proportional to the actual values,
the experimental results should reflect the trend of the actual case.}
The values of these ratios to the real numbers are shown in Table~\ref{tab:sim_ratio}.

\begin{table*}
\small
\centering
\caption{Operation costs and shelter based on a previous work \cite{otsuka2016}.
Unit of $f_m$ is dollars per month.
To make the calculation feasible,
100 shelters were distributed in proportion to the number from a previous work ~\cite{kumagai1995}.
Shelter capacities were set proportional to the average area of each type of facility.
}
\begin{tabular}{c|rrr|rrrrrrrrrr}
\toprule
 & \multicolumn{3}{c|}{Common}& \multicolumn{10}{c}{Wards of Kobe City}\\ \hline
Types of facilities $l$  &   $f_m$ & $C_m$&$U_{\ell}$& (1) & (2) & (3) & (4) & (5) & (6) & (7) & (8) & (9) & Total\\
\midrule
Daycares/kindergartens & 11,100 &3 &4,353& 1 & 1 & 1 & 1 & 1 & 1 & 1 & 0 & 0 & 7 \\
Elementary schools     & 54,900 &23& 19,882& 3 & 2 & 2 & 2 & 3 & 3 & 3 & 1 & 2 & 21 \\
Junior high schools     & 52,800 &22& 7,437& 1 & 1 & 1 & 1 & 1 & 1 & 1 & 0 & 1 & 8 \\
High schools      & 66,600 &24& 4,652& 1 & 0 & 0 & 0 & 1 & 1 & 1 & 0 & 0 & 4 \\
Universities/colleges     & 36,900 &21& 788& 1 & 0 & 0 & 0 & 0 & 0 & 0 & 0 & 0 & 1 \\
Public facilities (small) & 13,500 &2& 14,866& 4 & 3 & 3 & 4 & 2 & 1 & 1 & 0 & 1 & 19 \\
Public facilities (medium) & 57,300 &3& 9,406& 2 & 2 & 1 & 2 & 1 & 0 & 0 & 0 & 1 & 9 \\
Public facilities (large) & 89,400 &8& 2,749& 1 & 1 & 0 & 1 & 0 & 0 & 0 & 0 & 0 & 3 \\
Private facilities (small) & 27,900 &2& 9,783& 0 & 4 & 6 & 5 & 1 & 3 & 0 & 0 & 0 & 19 \\
Private facilities (large) & 137,400 &4& 388& 0 & 0 & 1 & 0 & 0 & 0 & 0 & 0 & 0 & 1 \\
Parks      & 2,100   &6& 7,736& 1 & 2 & 2 & 1 & 2 & 0 & 0 & 0 & 0 & 8 \\
\midrule
Total      & - & -& 82,040& 15 & 16 & 17 & 17 & 12 & 10 & 7 & 1 & 5 & 100 \\ 
\midrule
\multicolumn{4}{c|}{Area (km$^2$)} &  34 & 33 & 29 & 15 & 11 & 29 & 28 & 138 & 240 & 552\\
\bottomrule
\end{tabular}
\label{tab:shelters}
\end{table*}

\begin{table*}[t]
\small
\centering
\caption{{\sc Hanshin} setting based on 
number of evacuation shelters and people in them during Kobe earthquake~\cite{kumagai1995,KobeMinsei1996}
}
\begin{tabular}{rrr|rrrrrrrrrrr}
\toprule
 \multicolumn{3}{c|}{Kobe earthquake~\cite{KobeMinsei1996}} & \multicolumn{11}{c}{{\sc Hanshin} $|N_t|$} \\
Date       & \# shelters & \# evacuees & $t$ & (1) & (2) & (3) & (4) & (5) & (6) & (7) & (8) & (9) & Total \\
\midrule
1995/1/17 & 497  & 202,043 & 0 &296 & 132 & 142 & 112 & 197 & 88 & 15 & 4 & 14 & 1000\\
1995/2/17 & 527  & 106,050 & 1 &155 & 69 & 75 & 59 & 103 & 47 & 8 & 2 & 8 & 526\\
1995/3/17 & 442  & 62,604  & 2 &91 & 40 & 44 & 35 & 61 & 28 & 5 & 1 & 5 & 310\\
1995/4/17 & 391  & 42,330  & 3 &61 & 27 & 30 & 24 & 41 & 19 & 4 & 0 & 4 & 210\\
1995/5/17 & 361  & 31,132  & 4 &45 & 20 & 22 & 18 & 30 & 14 & 3 & 0 & 3 & 155\\
1995/6/17 & 314  & 21,609  & 5 &31 & 14 & 15 & 13 & 21 & 10 & 2 & 0 & 2 & 108\\
1995/7/17 & 283  & 16,748  & 6 &24 & 11 & 12 & 10 & 16 & 8 & 2 & 0 & 2 & 85\\
1995/8/17 & 222  & 8,491   & 7 &12 & 6 & 6 & 5 & 8 & 4 & 1 & 0 & 1 & 43\\
\bottomrule
\end{tabular}
\label{tab:evacuee}
\end{table*}

\begin{table}
\small
\centering
\caption{Ratio of simulation setting of dataset {\sc Hanshin}
to original data~\cite{KobeMinsei1996}
}
\begin{tabular}{l|r|rr}
\toprule
    & Original   &  {\sc Hanshin} & Ratio $r$ \\ \midrule
Evacuees & 202,043 & 1,000 & 202.0  \\
Shelters & 693    & 100 & 6.93  \\
\bottomrule
\end{tabular}
\label{tab:sim_ratio}
\end{table}

The shelter capacities in Table~\ref{tab:shelters} were
set to fill 90\% of the capacity 
if all the evacuees were accommodated at $t=1$.
$\bm{p}_m$ were plotted randomly
on a square of the same area as the ward.
The return time and initial location of an evacuee are set to be independent for simplicity, 
although they may not be independent because the disaster situation obviously depends on the region.
The movement cost was set according to Eqs.~(\ref{eq:move0}) and~(\ref{eq:move1}), and $\alpha$ was set to $10$.
Ten datasets were generated for both training and test.

\subsection{Comparing methods}

We compared the proposed method with those that have the following procedures in place of {\sc Opt}:

\begin{itemize}

\item Sequential FLP: {\sc SeqFlp} 

We used {\sc SeqFlp} in the proposed method as the baseline.

\item No Move Model: {\sc NoMove}

At $t=0$, the evacuees are accommodated in the shelters to minimize
the sum of the evacuation costs.
At $t>0$, the evacuees do not relocate; they stay in the same shelter as $t=0$
until they return home.
This procedure minimizes the movement cost.

\item Bin-Packing Model: {\sc BinPack}

{\sc BinPack} ignores the movement costs
and solved the following bin-packing problem
to get $y_{tm}$:
\begin{align}
& \text{Minimize} & & \sum_{t=0}^{T-1} \sum_{m \in M} f_{m} y_{tm} \tag{\ref{eq:obj}.y}\\ 
& \text{Subject to} & & |N_t| \leq \sum_{m \in M} C_m y_{tm}   \tag{\ref{eq:capa}.y} \\ 
& & & y_{(t+1)m} \leq y_{tm}  \tag{\ref{eq:noreopen}.y} \\ %
& & & y_{tm} \in \{ 0,1 \}. \tag{\ref{eq:xrange}.y} 
\end{align}

Then we solved the following optimization problem with $y_{tm}$ fixed
and determined $x_{tmn}$. 
This method yields a solution that minimizes the operation cost of the shelters:

\begin{align}
& \text{Minimize} & & \sum_{t=0}^{T-1} \Big( \sum_{m \in M} \sum_{m' \in M} d_{tmm'} \sum_{n \in N} z_{tmm'n} \Big) \tag{\ref{eq:obj}.x} \\ 
& \text{Subject to} & & \sum_{n \in N} x_{tmn} \leq C_m y_{tm} ,  \forall m \tag{\ref{eq:capa}.x} \\ 
& & & \sum_{m \in M} x_{tmn} = 1, \forall t \leq \tau_n \tag{\ref{eq:alive}.x} \\ 
& & & \sum_{m \in M} x_{tmn} = 0, \forall t > \tau_n \tag{\ref{eq:dead}.x} \\ %
& & & z_{tmm'n} \geq x_{tmn} + x_{(t+1)m'n} - 1  \tag{\ref{eq:z1}.x} \\ %
& & & z_{tmm'n} \leq x_{tmn}  \tag{\ref{eq:z2}.x} \\ %
& & & z_{tmm'n} \leq x_{(t+1)m'n}  \tag{\ref{eq:z3}.x} \\ %
& & & x_{t=0, mn} = \begin{cases}
 1, &  m = \tilde{m}_0(n)\\
 0, & m \neq \tilde{m}_0(n)
 \end{cases} \tag{\ref{eq:init}.x} \\ %
& & & x_{tmn}, z_{tmm'n} \in \{ 0,1 \}  \tag{\ref{eq:xrange}.x} %
\end{align}


\RED{
\item Binary Quadratic Problem: {\sc Bqp}

{\sc Bqp} is a procedure for solving Eqs.~\eqref{eq:obj}~--~\eqref{eq:xrange} as a quadratic programming problem.
}

\end{itemize}

\subsection{Experimental environment}

The experiments in this paper were performed on a computer with an Intel(R) Core (TM)i7-1065G7, 1.50GHz CPU, and 16GB memory.
Gurobi~\cite{gurobi} was used as an integer linear programming solver.

\section{Results}

\subsection{Estimation of movement cost} 
\label{sec:result_estimate}

\begin{figure}
\includegraphics[width=7cm]{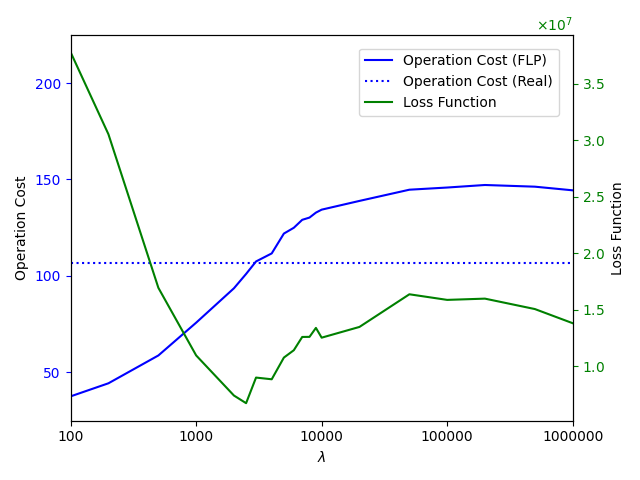}
\caption{
Estimation of movement cost ($\lambda$) 
with dataset {\sc Hanshin Train}.
}
\label{fig:estimate}
\end{figure}

The loss function (Eq.~\ref{eq:MSE}) values for different $\lambda$
are shown \RRED{by the green line} in Fig.~\ref{fig:estimate}. 
\RRED{
}
When $\lambda=2,500$, the loss function
was minimized
and best fit to the data.
In addition to the loss function,
we compare the total operation cost ($106.6$ million dollars~\cite{otsuka2016}, \RRED{shown by the dotted blue line in Fig.~\ref{fig:estimate}})
 with the estimated operation cost
\RRED{ (the blue line)}:
\begin{align}
r_m \times \sum_{t=0}^{T-1} \sum_{m \in M} y_{tm} f_{m}.
\label{eq:ope_cost}
\end{align}
At $\hat{\lambda}=2,500$,
the corresponding estimated operation cost was
also close to the actual value: 101.1 million dollars
(Fig.~\ref{fig:estimate}).
On the other hand, for the same dataset {\sc Hanshin Train}, applying {\sc NoMove} caused an excessive operation cost of 146.7 million dollars, which overestimated the reality,
 and applying {\sc BinPack} resulted in an insufficient operation cost of 25.4 million dollars, which underestimated the reality.

Based on the above results, we assume that in the Kobe earthquake, evacuation shelters were operated using a method such as {\sc SeqFlp}.
\RED{By dividing $\hat{\lambda}$ by the ratio $r_n$,
we got 1,240 dollars per person per km for the evacuation cost,
and 124 dollars per person per km for the relocation cost.}
This relocation cost resemble required nuisance fees for the relocation of evacuees.

\subsection{Performance Evaluation}
\label{sec:performance}
\begin{table*}[tb]
\small
\centering
\caption{
Objective function of simulation for dataset {\sc Hanshin Test} and its breakdown of movement and operation costs.
Total number of times evacuees moved is shown with movement cost.
{\bf Bold} type indicates the smallest value in each line.
All values are averages of ten trials.
The calculation time per trial is shown in the bottom line.
Parameter $\hat{\lambda}$ was fixed to $2,500$.
\RED{
Note that the relocation cost is not proportional to the number of times, but to the distance.
}
}
\begin{tabular}{r|rrrrr}
\toprule
Methods    & {\sc SeqFlp}   & {\sc NoMove} & {\sc BinPack} & \RED{\sc Bqp} & {\sc Opt} \\ \midrule
Objective ($\times 10^3$) & 27,393.7  & 27,963.2  & 37,870.9 & 27,370.3 & {\bf 26,937.4}  \\   \hline
Evacuation cost ($\times 10^3$)& 25,839.4  & {\bf 25,836.0}  & 36,825.0  & 25,862.6 & 25,870.8  \\ \hline
Relocation cost ($\times 10^3$) & 55.2      & {\bf 0.0}   & 679.7 & 578.3 & 53.3      \\ 
\RED{$\sum_{t=1}^T \sum_{m} \sum_{m' \neq m} \sum_{n} z_{tmm'n}$} (times) & 28.7  & {\bf 0.0}   & 115.8  & 151.1       & 32.6  \\  \hline
Operation cost ($\times 10^3$) & 1,499.1   & 2,127.2   & {\bf 366.2}& 929.4 & 1,013.3   \\ 
\RED{Eq.~(\ref{eq:ope_cost})} (million dollars) &  103.9         & 147.4         & {\bf 25.4} & 64.4 & 70.2  \\ \hline
Computational time (s) & 4.9       & 1.7       & 84.4 & 1,652.6 & 510.8  \\ 
\bottomrule
\end{tabular}
\label{table:hanshin_test}
\end{table*}

The experimental results of the performance evaluation using the dataset {\sc Hanshin Test} are shown in Table~\ref{table:hanshin_test}.
The movement cost was fixed at $\hat{\lambda}=2,500$.
In all the methods, the evacuation cost accounts for most of the objective function,
and the relocation and operation costs are less than one-tenth of the evacuation cost.
This is because the evacuation cost was set ten times higher than the relocation cost ($\alpha=10$)
to emphasize the importance of quickly reaching a shelter during a disaster.
Compared to the baseline {\sc SeqFlp}, {\sc NoMove} lowered the movement costs, although it 
significantly increased the operation costs.
{\sc BinPack} also decreased the operation costs, but it increased the movement costs.
On the other hand, {\sc Opt} minimizes the objective function.
Although the operation cost of {\sc Opt} exceeds {\sc BinPack},
the number of relocations of {\sc Opt} is less than {\sc BinPack} 
and close to {\sc SeqFlp}.
The operation cost of {\sc Opt} was 70.2 million dollars,
while that of {\sc SeqFlp} was 103.9 million dollars.
The proposed method would have reduced the operation cost of the 
shelters in the Kobe earthquake
by 33.7 million dollars (32\%) 
if the return time of the evacuees had been known in advance.
\RED{
By comparing {\sc Opt} with {\sc Bqp}, it was confirmed that converting the quadratic objective function to a linear one was effective in terms of both computation time and performance.
}

\begin{figure}[tb]
\center
\small
\includegraphics[width=7cm]{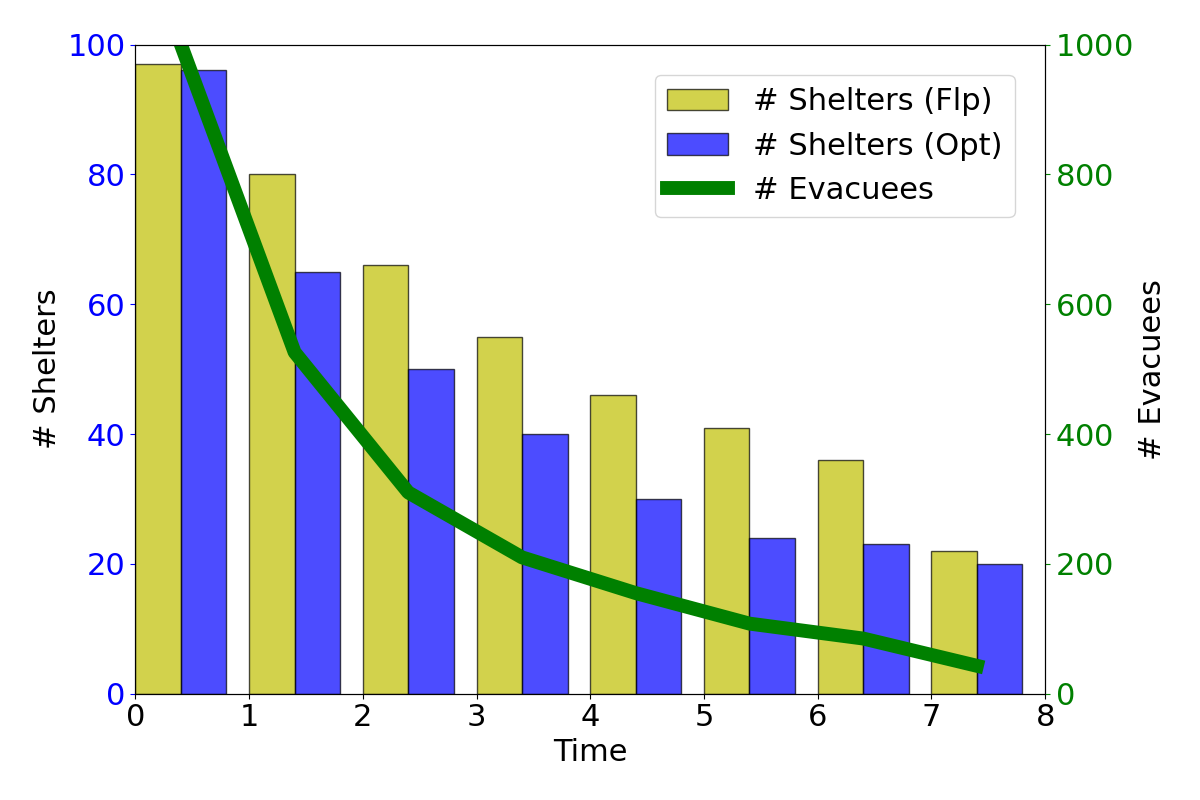}
\caption{
History of number of evacuees and shelters.
Horizontal axis shows time $t$,
left vertical axis shows operating shelters,
and right vertical axis shows number of evacuees.
All values are averages of ten trials.
}
\label{fig:history}
\end{figure}

To compare the reduction in operation costs in detail,
the histories of the number of shelters operated by {\sc SeqFlp} and {\sc Opt}
are shown in Fig.~\ref{fig:history}.
\RED{
Since evacuation cost was set to be ten times higher than relocation cost,
evacuation cost is given more importance than the operation cost at $t=0$,
so there is little difference between the two methods.
While the number of shelters decreased as the number of evacuees decreased,
the rate of decrease was less than the number of evacuees.
This is because at least one shelter is required even for a small number of evacuees,
and the smaller the number of evacuees, the lower the efficiency.
However, since the number of shelters was reduced
more quickly in {\sc Opt} than in {\sc SeqFlp},
and there is a large difference during $t=1, \cdots, 6$.
This is the breakdown of the operation cost reduction by the proposed method.
}

\begin{figure}[tb]
\centering
\includegraphics[width=7cm]{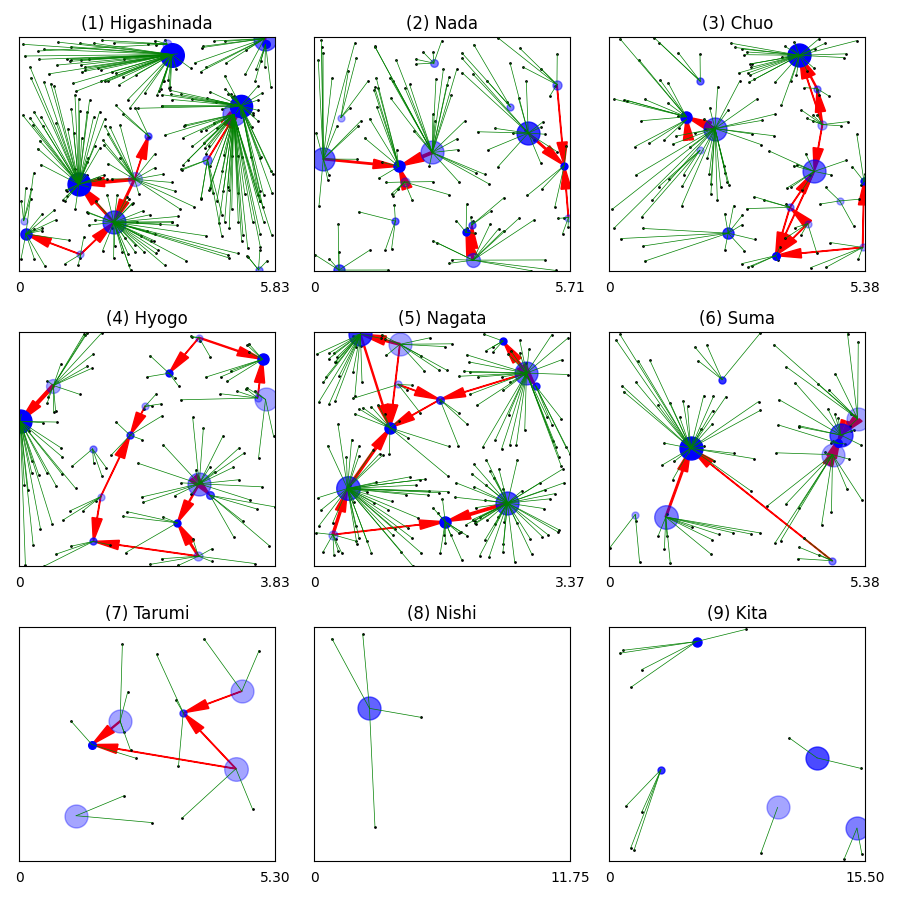}
\label{fig:Trajectory_Opt}
\caption{
Trajectory of evacuees in one dataset {\sc Hanshin Test}.
Movement cost was fixed at $\hat{\lambda}=2,500$.
}
\label{fig:Trajectory}
\end{figure}

\RED{
The trajectories of the evacuees for the proposed method are shown in Fig.~\ref{fig:Trajectory}.
\RRED{
Blue circles represent shelters.
The size of blue circles represents its capacity,
and the darker blue denotes its longer operation period.
Small black dots represent evacuees,
and green lines represent their evacuation trajectories to first shelter.
Red lines represent relocations between shelters,
and thickness of lines represents number of relocated people.
Each of the nine wards of Kobe City corresponds to a square with equal area.
The length of one side is shown 
 in kilometer at the bottom right of the square.
}
Comparing the red arrows between the shelter of origin and the shelter of destination, we can see that in the optimal solution,
the evacuees are often relocated to shelters with smaller capacity.
In general,
long-term stays in evacuation shelters require bulky equipment for living, which increases the operation cost.
However, these experimental results suggest that an effective countermeasure is to deploy some shelters that can be operated at low cost for a long-term period in an easily accessible place from surrounding shelters, even if their capacities are small.
}


\section{Conclusion \& Future Work}

We formulated an evacuation shelter scheduling problem
that takes into account the closure of shelters due to the sequential return of evacuees to their homes.
We also proposed a method that estimates the movement cost of evacuees.
Our simulation experiments on the Kobe earthquake
 showed that our proposed method reduced operation costs by 33.7 million dollars: 32\%.
\RED{
Note that the proposed method does not guarantee to reduce all of the evacuation, relocation, and operation costs,
but to minimize the total cost.}
Although we only considered a decrease in the number of evacuees in the formulation,
a natural extension could also address an increase in them.

We conclude by describing three issues as future work.
\RED{
The first one is to incorporate elements into our model
such as the modalities of transportation for evacuation~\cite{swamy2017hurricane},
the urgency of the initial evacuation,
and special facilities for pets and the elderly~\cite{kocatepe2018pet},
depending on the target disaster.
}
The second one will develop a more efficient calculation method
for larger $|N|$, $|M|$, and $T$.
\RED{
The third issue is to obtain when evacuees return to their homes,
for example, by predicting when transportation will be restored
and when temporary housing will be built.
Although such return times of evacuees are assumed to be given,
this issue will be critical for the application of the proposed method.
}




\bibliographystyle{ieeetr}
\bibliography{hicss}

\end{document}